\documentclass[letterpaper]{article} % DO NOT CHANGE THIS
\usepackage{aaai2026}  % DO NOT CHANGE THIS
\usepackage{times}  % DO NOT CHANGE THIS
\usepackage{helvet}  % DO NOT CHANGE THIS
\usepackage{courier}  % DO NOT CHANGE THIS
\usepackage[hyphens]{url}  % DO NOT CHANGE THIS
\usepackage{graphicx} % DO NOT CHANGE THIS
\usepackage{natbib}  % DO NOT CHANGE THIS AND DO NOT ADD ANY OPTIONS TO IT
\usepackage{caption} % DO NOT CHANGE THIS AND DO NOT ADD ANY OPTIONS TO IT
\usepackage{float}
\usepackage{placeins}
\usepackage{algorithm}
\usepackage{algorithmic}
\usepackage{newfloat}
\usepackage{listings}
\usepackage{booktabs}
\usepackage{amsmath}   % 提供 aligned / align / split 等环境
\usepackage{amssymb}
\usepackage{newfloat}
\usepackage{listings}
\usepackage{multirow}
\usepackage{booktabs}
\usepackage{adjustbox}
\usepackage[table]{xcolor}
\DeclareCaptionStyle{ruled}{labelfont=normalfont,labelsep=colon,strut=off} % DO NOT CHANGE THIS
\floatstyle{ruled}
\newfloat{listing}{tb}{lst}{}
\floatname{listing}{Listing}
\title{Semantics and Content Matter: Towards Multi-Prior Hierarchical  Mamba \\for Image Deraining}
\author {
    Zhaocheng Yu\textsuperscript{\rm 1},
    Kui Jiang\thanks{Corresponding Author}\textsuperscript{\rm 1},
    Junjun Jiang\textsuperscript{\rm 1},
    Xianming Liu\textsuperscript{\rm 1},
    Guanglu Sun\textsuperscript{\rm 2},
    Yi Xiao\textsuperscript{\rm 3}
}
\affiliations {
    \textsuperscript{\rm 1}Faculty of Computing, Harbin Institute of Technology\\
    \textsuperscript{\rm 2}School of Computer Science and Technology, Harbin University of Science and Technology\\
    \textsuperscript{\rm 3}School of Computer and Artificial Intelligence, Zhengzhou University\\
    yuzhaocheng@stu.hit.edu.cn, \{jiangkui,jiangjunjun,csxm\}@hit.edu.cn, 
    Sunguanglu@hrbust.edu.cn,
    yixiao@zzu.edu.cn
}
\usepackage{bibentry}
\begin{document}

\maketitle

\begin{abstract}
Rain significantly degrades the performance of computer vision systems, particularly in applications like autonomous driving and video surveillance. 
While existing deraining methods have made considerable progress, they often struggle with fidelity of semantic and spatial details.  
To address these limitations, we propose the Multi-Prior Hierarchical Mamba (MPHM) network for image deraining. This novel architecture synergistically integrates macro-semantic textual priors (CLIP) for task-level semantic guidance and micro-structural visual priors (DINOv2) for scene-aware structural information. To alleviate potential conflicts between heterogeneous priors, we devise a progressive Priors Fusion Injection (PFI) that strategically injects complementary cues at different decoder levels. 
Meanwhile, we equip the backbone network with an elaborate Hierarchical Mamba Module (HMM) to facilitate robust feature representation, featuring a Fourier-enhanced dual-path design that concurrently addresses global context modeling and local detail recovery. 
Comprehensive experiments demonstrate MPHM's state-of-the-art performance, achieving a 0.57 dB PSNR gain on the Rain200H dataset while delivering superior generalization on real-world rainy scenarios. %with high-fidelity restoration.
% Comprehensive experiments across diverse benchmarks demonstrate that MPHM achieves a new state-of-the-art performance, delivering an average gain of 0.15 dB PSNR on five synthetic datasets while generalizing robustly to real-world data with superior restoration quality.

% Comprehensive experiments across diverse synthetic and real-world benchmarks confirm that MPHM %not only establishes 
% achieves new state-of-the-art performance with low computational overhead. % demonstrating its unique capability to deliver high-quality, semantically consistent deraining results while maintaining practical efficiency.
\end{abstract}


\section{Introduction}
Adverse weather conditions degrade image quality, impairing the performance of advanced computer vision algorithms. This poses significant challenges in critical applications, such as autonomous vehicles and surveillance systems~\cite{dang2024adaptive, dang2023efficient,hong2025resilience}.
\begin{figure}[h]
\centering
\includegraphics[width=\linewidth]{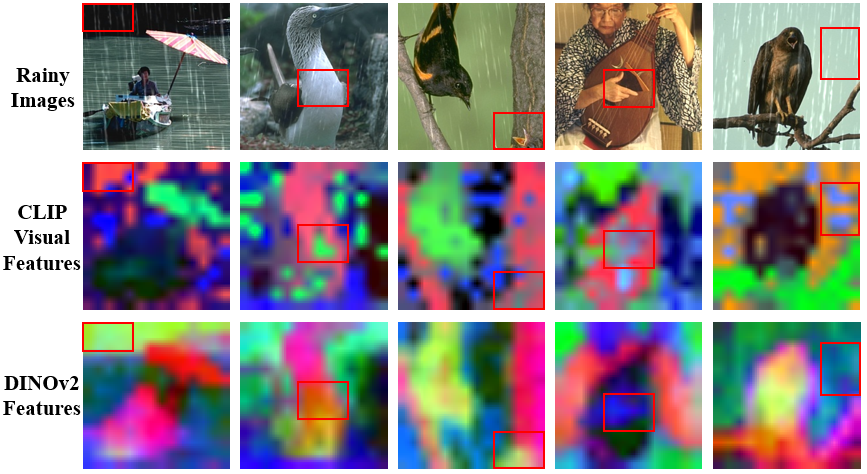}
\caption{For analytical purposes, this visualization displays the PCA (Principal Component Analysis) projection of high-dimensional features from the visual encoders of various pre-trained models into 3-dimensional space.}
\vspace{-4mm}
\label{fig:motivation}
\end{figure}

The image deraining task aims to restore high-quality visual content from rain-affected inputs and is a fundamental component of modern computer vision systems. Traditional methods~\cite{kang2011automatic,chen2013generalized} typically rely on rain pattern analysis and handcrafted priors for rain removal. However, their performance is often constrained by strong scene-specific dependencies, resulting in poor generalization across diverse real-world applications~\cite{zhong2022rainy}.

Recent advances in deep learning~\cite{lecun2015deep} have significantly improved image deraining. CNN-based methods~\cite{wang2020dcsfn, guo2023sky, ren2019progressive} excel at recovering local textures through hierarchical convolutions but are constrained by limited receptive fields, hindering their ability to capture long-range dependencies. Transformer-based~\cite{xiao2022image, chen2021pre} and Mamba-based~\cite{li2024fouriermamba} approaches overcome this limitation by %employing self-attention to model global context, 
aggregating global response with self-attention or state space module (SSM), leading to substantial performance gain. However, these methods struggle with either insufficient semantic completeness or limited spatial detail preservation.
%better restoration quality. %However, their quadratic computational complexity results in high costs, especially for high-resolution images.

% Mamba~\cite{liu2024vmamba} offers an effective solution to these challenges, combining linear computational complexity with superior global modeling capabilities. 
% ~\cite{sun2024hybrid, yamashita2024image} proposed a U-shaped multi-scale architecture that connects Mamba blocks through skip connections, enabling effective multi-scale feature extraction and improved deraining performance.
% However, current Mamba-based deraining networks still face several limitations:
% \textit{i}) converting spatial tensors to 1D sequences disrupts local spatial relationships in scene representation;
% \textit{ii}) %The model's performance remains suboptimal in practice due to 
% insufficient semantic understanding when processing complex real-world scenes with diverse rain patterns.

Recent breakthroughs in large models have generated significant interest in their scene perception capabilities~\cite{radford2021learning,oquab2023dinov2}. These models typically employ an architecture where images are converted by a visual encoder into a series of visual tokens or by a text encoder into language-aligned feature descriptions~\cite{bai2025textir}.  Inspired by this, large-model priors have also been applied to image restoration tasks, such as extracting the macro-level degradation description~\cite{jin2024llmra} or micro-level background textures~\cite{lin2023multi}, yielding substantial performance gains. 
% To overcome the semantic understanding limitations of deraining models, researchers have increasingly turned to pre-trained large models due to their superior scene generalization capabilities.
% For instance, ~\cite{jin2024llmra} employs the CLIP ~\cite{radford2021learning} text encoder to generate macro-level degradation priors. 
% These priors are then injected into backbone features, providing task-specific semantic guidance.
%From a different perspective, ~\cite{lin2023multi} introduces a Mixture-of-Experts (MoE) module to fuse multi-level features extracted from the pre-trained DINOv2 ~\cite{oquab2023dinov2} model.  This approach leverages micro-level scene-aware priors, significantly enhancing texture reconstruction.
% Although prior-based guidance is effective, using either macro-level textual priors or micro-level visual priors alone fails to achieve an optimal balance between semantic understanding and structural precision. 
Image deraining tasks encompass various rain patterns with distinct characteristics. This diversity makes it challenging for a single prior—whether textual or visual—to optimally balance semantic completeness and content fidelity across all scenarios. 
To address this, ~\cite{luo2023controlling} integrate CLIP's textual and visual encoders~\cite{radford2021learning} into a unified framework to provide comprehensive semantic guidance for image restoration. 
However, CLIP's visual encoder prioritizes global image-text alignment, favoring abstract semantics over detail. 
As shown in Figure~\ref{fig:motivation}, it is also sensitive to noise, producing suboptimal features for rainy scenes. This phenomenon may be due to CLIP’s training data and objective, where the global-level image-text contrastive learning makes the image encoder poor at capturing visual details. 
In contrast, the DINOv2 encoder~\cite{oquab2023dinov2} generates finer-grained representations with richer semantic content. 
This suggests a promising approach: harmonizing DINOv2's micro visual priors with CLIP's macro textual priors to adapt to diverse visual characteristics and requirements. 
Unfortunately, the independent architectures of CLIP and DINOv2 prevent joint optimization with multimodal data. Consequently, each encoder remains confined to its original pre-trained capabilities, limiting their potential to leverage synergistic advantages in multimodal representation.
To this end, we develop the Multi-Prior Hierarchical Mamba (MPHM) framework %, a multi-prior joint learning framework 
to complement semantics and details while alleviating modality conflicts encountered by multiple encoders for high-accuracy image deraining. 
%\textit{i}) 
To promote semantic completeness, 
%we devise a multi-prior fusion framework where
MPHM sequentially integrates the scene-aware visual priors (micro) generated by DINOv2 and the task-level textual priors (macro) from CLIP into all decoder stages of the backbone network.
This sequential Priors Fusion Injection (PFI) strategy significantly alleviates feature conflicts that may arise from direct fusion of multi-prior representations, while ensuring comprehensive semantic guidance throughout the network. 
% \textit{ii}) 
To enhance texture representation, we introduce a Hierarchical Mamba Module (HMM) that enables multi-level global-local interactions and cross-channel knowledge integration. 
This design effectively addresses Mamba's limitations in capturing local spatial relationships. 
Furthermore, through combined spatial and frequency domain processing, HMM establishes robust structural and textural representations.
In general, the contributions of this work can be summarized as follows.
\begin{itemize}
    \item We propose a novel Multi-Prior Hierarchical Mamba (MPHM) framework that systematically integrates complementary semantic and detail priors from multiple pre-trained foundation models to achieve robust image deraining.
    \item We introduce %a multi-stage prior fusion scheme 
    a Priors Fusion Injection (PFI) where DINOv2's visual priors and CLIP's textual priors are progressively fused across decoder levels to jointly enhance spatial detail and semantic representation. A Hierarchical Mamba Module (HMM) is proposed to augment Mamba's local spatial modeling capability through multi-level global-local feature interactions, thereby improving structural representation and restoration robustness via integrated spatial-frequency processing.
    \item Extensive experiments on both synthetic and real-world datasets demonstrate that our proposed MPHM method consistently outperforms state-of-the-art approaches in terms of both quantitative metrics and qualitative results.
\end{itemize}

\section{Method}
% This section details the technical framework of the proposed method. First, we provide an overview of the system workflow and optimization losses. Next, we comprehensively describe the key components, including the network architecture.
This section details the proposed framework. We first outline the system workflow and optimization objectives and then comprehensively describe the key components.

\subsection{Pipeline Overview}
\begin{figure*}[htbp]
  \centering
  \includegraphics[width=0.95\textwidth]{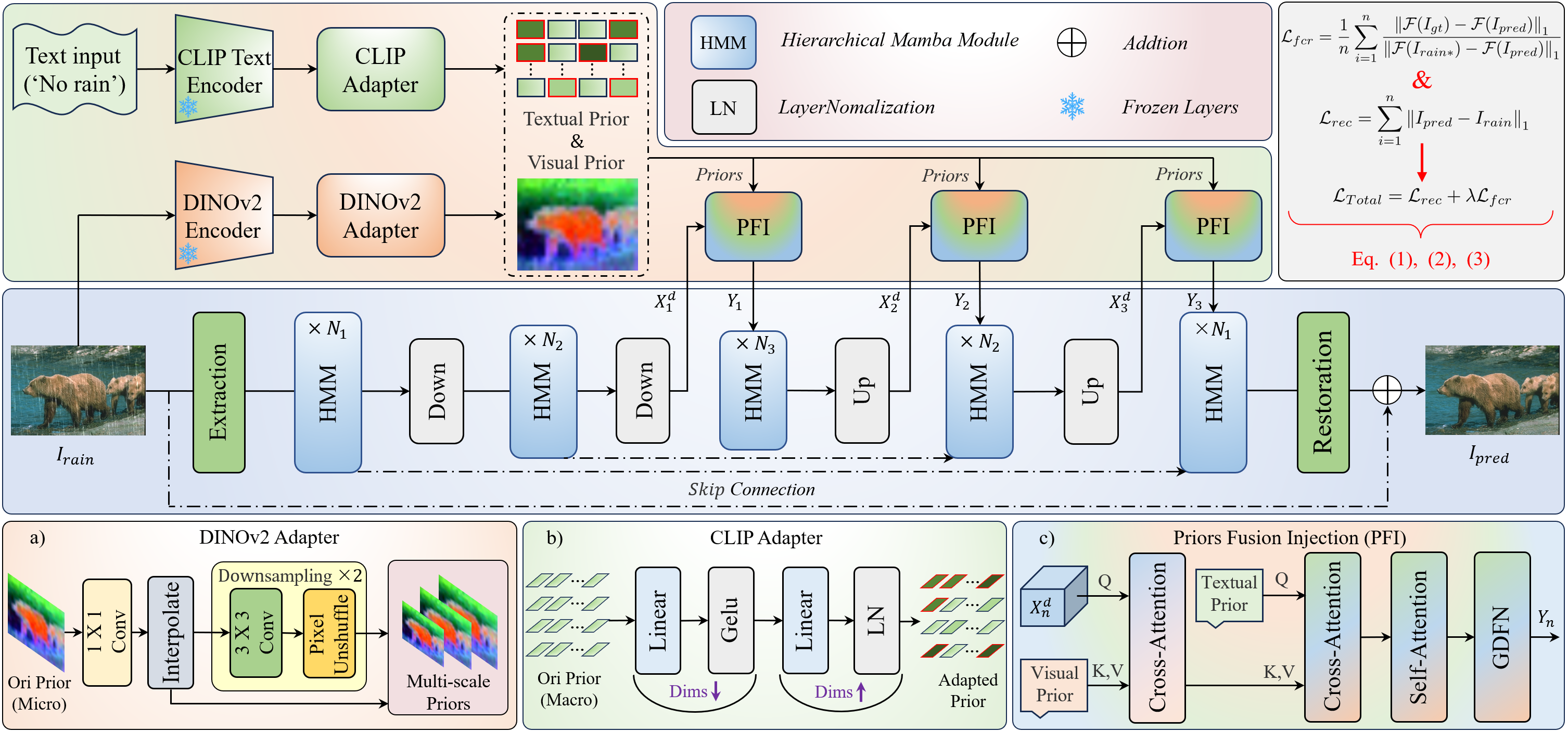}\vspace{-2mm}
  \caption{Overview of the proposed Multi-Prior Hierarchical Mamba deraining framework. (a) DINOv2 Adapter, (b) CLIP Adapter and (c) Priors Fusion Injection (PFI).} \vspace{-4mm}
  \label{pipeline}

\end{figure*}

% \subsubsection{Architecture Overview.} 
Figure~\ref{pipeline} illustrates the MPHM architecture, comprising two core components: a U-shaped encoder-decoder backbone built with Hierarchical Mamba Modules (HMM), and a prior generation-injection pipeline. The backbone progressively extracts and refines dual-domain (spatial/frequency) features from rainy input $I_{\text{rain}}$, with skip connections that maintain structural consistency.
%Figure~\ref{pipeline} illustrates the overall architecture of the proposed MPHM network, which consists of two key parts: a backbone deraining network and a prior generation-injection pipeline.
%The backbone adopts a U-shaped encoder–decoder design constructed with Hierarchical Mamba Module (HMM), which progressively extracts and refines dual-domain (spatial and frequency) features from the rainy image $I_{rain}$. Skip connections preserve structural consistency across processing stages.
The prior generation system provides complementary semantic guidance. We generate the scene-level visual prior $P_v$ from $I_{\text{rain}}$ using a frozen DINOv2 encoder and lightweight adapter, while deriving the task-level textual prior $P_t$ from handcrafted description (``No rain") via the frozen CLIP text encoder and refinement adapter.
%In parallel, the prior generation system introduces two complementary sources of semantic guidance. The scene-level visual prior $P_{v}$ is derived from the rainy image $I_{rain}$ via a frozen DINOv2 encoder followed by a lightweight adapter that generates multi-scale structural features. The task-level textual prior $P_{t}$ is obtained from a handcrafted description (No rain) using a frozen CLIP text encoder and is then refined by a CLIP adapter to enhance task relevance.
% To bridge these priors with the backbone network, we design the Prior Fusion Injection (PFI) module, which is inserted at each decoding stage. 
% PFI employs a dual-stage cross-attention mechanism to inject the texture and task priors into the backbone features, followed by self-attention and a gated feedforward block (GDFN) for semantic propagation and refinement.
% The final derained image $I_{pred}$ is reconstructed by adding the predicted rain residual to the input $I_{rain}$, thus completing the restoration process.
These priors are integrated at each decoder stage via Priors Fusion Injection (PFI) modules. PFI employs a dual-stage cross-attention mechanism to inject texture and task priors into the backbone features, followed by self-attention and a gated feedforward block (GDFN) for semantic propagation and refinement. 
The final derained image $I_{\text{pred}}$ is reconstructed by eliminating the predicted rain residual from $I_{\text{rain}}$.

% \subsubsection{Model Optimization.} To maintain visual fidelity while preserving high-frequency textures of the restored images, we implement the $L1$ loss and frequency-adaptive contrastive loss $L_{fcr}$, mathematically defined as follows: 
% \begin{equation}
% \mathcal{L}_{rec} = \sum_{i=1}^{n} \left\| I_{pred} - I_{rain} \right\|_1,
% \end{equation}
% \begin{equation}
% \mathcal{L}_{fcr} = \frac{1}{n} \sum_{i=1}^{n} \frac{\left\|\mathcal{F}(I_{gt}) - \mathcal{F}(I_{pred}) \right\|_1}{\left\| \mathcal{F}(I_{rain*}) - \mathcal{F}(I_{pred}) \right\|_1},
% \end{equation}

% where $I_{gt}$ and $I_{\text{pred}}$ denote the corresponding ground truth and predicted rain-free image, while $I_{rain*}$ represents the input rainy image $I_{rain}$ and other rainy samples in the training batch. Finally, the equation of overall loss is
% \begin{equation}
% \mathcal{L}_{Total} =  \mathcal{L}_{rec} + \lambda \mathcal{L}_{fcr},
% \end{equation}
% where the value of $\lambda$ is $0.1$ to balance the loss weights.

\subsubsection{Model Optimization.}
To achieve faithful deraining and texture preservation, we minimize a hybrid loss that integrates pixel reconstruction with frequency-domain contrastive learning.
% For faithful deraining and texture preservation, we minimize a hybrid loss integrating pixel reconstruction and frequency-domain contrastive learning. 
The reconstruction loss ensures spatial accuracy:
% For faithful deraining and texture preservation, we minimize a hybrid loss combining pixel-level reconstruction with frequency-domain contrastive learning. The reconstruction loss is:
\begin{equation}
\mathcal{L}_{rec} = \sum_{i=1}^{n} \left\| I_{pred} - I_{gt} \right\|_1,
\end{equation}
where $I_{gt}$ and $I_{pred}$ denote the ground-truth and predicted rain-free images respectively. 
To enhance frequency-domain consistency, we incorporate the Frequency-domain Contrastive Regularization (FCR) loss~\cite{gao2024efficient} that minimizes the distance between derained spectra and clean references while maximizing their separation from rainy patterns.
% that pulls derained spectra to clean references while pushing away rainy patterns via contrastive learning.
% a contrastive learning method that aligns derained spectra with clean references while distancing them from rainy patterns.
The FCR loss is formulated as:
\begin{equation}
\mathcal{L}_{fcr} = \frac{1}{n} \sum_{i=1}^{n} \frac{\left\|\mathcal{F}(I_{gt}) - \mathcal{F}(I_{pred}) \right\|_1}{\left\| \mathcal{F}(I_{rain*}) - \mathcal{F}(I_{pred}) \right\|_1},
\end{equation}
where derained spectra $\mathcal{F}(I_{pred})$ are anchors, clean spectra $\mathcal{F}(I_{gt})$ are positives, and randomly sampled rainy spectra $\mathcal{F}(I_{rain*})$ are negatives. 
$\mathcal{F}(\cdot)$ operator denotes the Discrete Fourier Transform (DFT) and parameter $n = 2$ indicates the number of negative samples.
% The overall objective combines the pixel-level reconstruction term and the FCR term as:
The overall loss function is defined as:
\begin{equation}
\mathcal{L}_{Total} = \mathcal{L}_{rec} + \lambda \mathcal{L}_{fcr},
\end{equation}
where $\lambda=0.1$ is an empirically determined balancing weight that adjusts the trade-off between spatial-domain accuracy and frequency-domain fidelity.

\subsection{Hierarchical Mamba Module}

\begin{figure}[htbp]
  \centering
  \includegraphics[width=0.95\linewidth]{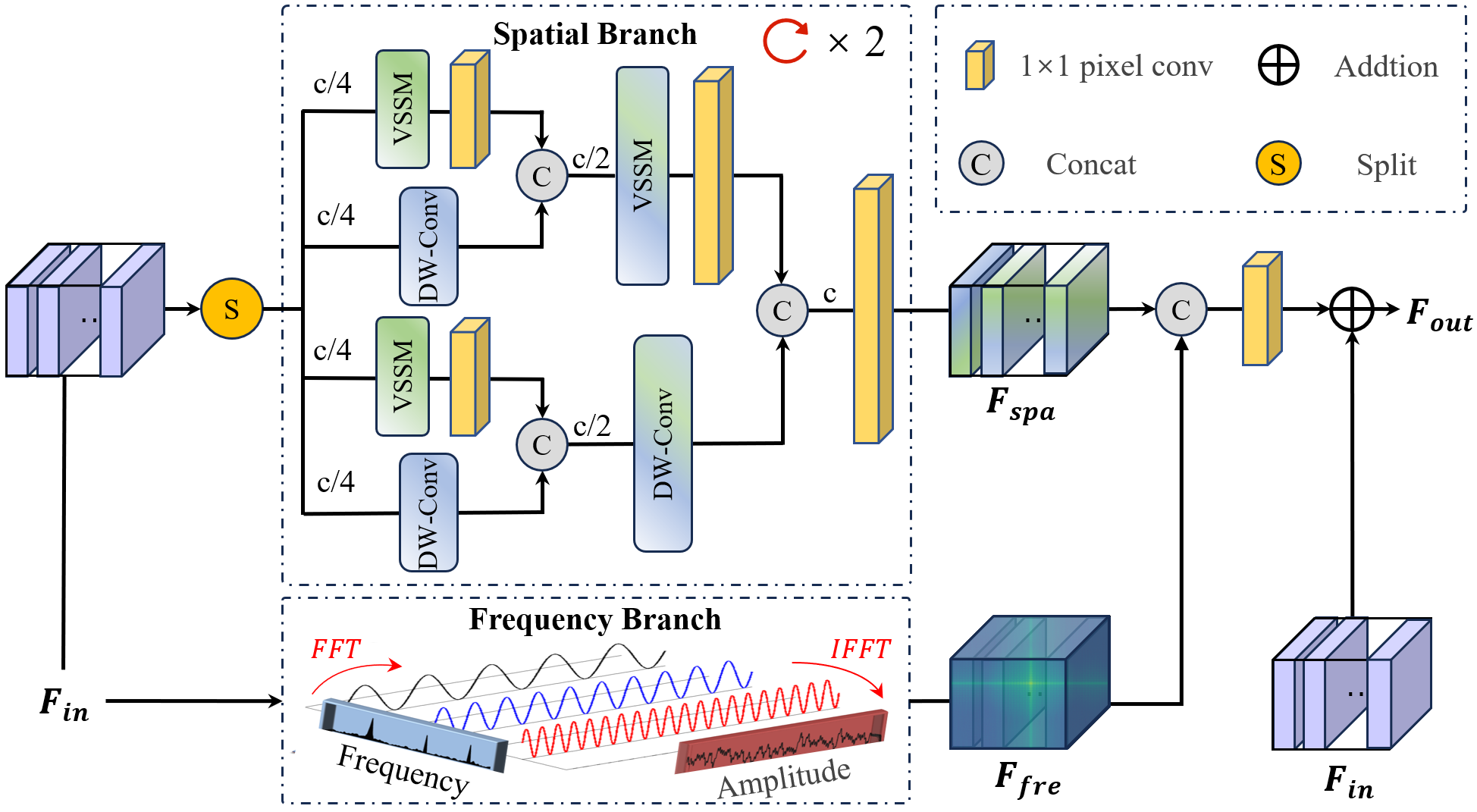}
  \caption{Pipeline of the HMM.}\vspace{-4mm}
  \label{SFM}
\end{figure}

%As shown in Figure~\ref{SFM}, HMM adopts a dual-branch parallel architecture that extracts spatial and frequency domain features from complementary perspectives. The specific structure of the above two branches is as follows:
As shown in Figure~\ref{SFM}, HMM processes features via parallel spatial and frequency branches. 
\subsubsection{Spatial-domain Branch.}
%To enhance the local spatial modeling capability of Mamba-based structures, we introduce a spatial branch that establishes hierarchical interactions global and local features across multiple levels.
To enhance local spatial modeling in Mamba-based architectures, we introduce a spatial branch that hierarchically integrates global and local features. 
Given input $F_{in} \in \mathbb{R}^{(b,c,h,w)}$, we first split features channel-wise into four equal partitions: $[f_1, f_2, f_3, f_4] = \mathcal{S}(F_{\text{in}})$ with $f_i \in \mathbb{R}^{(b,\frac{c}{4},h,w)}$. 
%we first divide it uniformly along the channel dimension into four parts, denoted as $F_{in} = [f_1, f_2, f_3, f_4]$, where each $F_{in} \in \mathbb{R}^{(b,\frac{C}{4},h,w)}$.
%These sub-features are processed in parallel to extract global and local cues through alternating Visual Selective Spatial Mamba (VSSM) and Depth-wise convolution (DConv) modules.
These sub-features undergo parallel processing through alternating modules. 
Specifically, $f_1$ and $f_3$ are passed through Visual Selective Spatial Mamba (VSSM)~\cite{liu2024vmamba} for global context modeling, while $f_2$ and $f_4$ are processed by depth-wise convolution (DW-Conv) to improve local feature continuity.
%All VSSM outputs subsequently pass through $1\times1$ convolutional layers for channel refinement at constant spatial resolution.
VSSM outputs are refined via $1\times1$ convolutions maintaining spatial resolution. 
We then concatenate the processed features pairwise (($f_1, f_2$) and ($f_3, f_4$))
%The outputs of ($f_1, f_2$) and ($f_3, f_4$) are concatenated respectively 
to form two intermediate features, denoted as $F_{GL1}$ and $F_{GL2}$.
%To strengthen the coupling between global and local representations, $F_{GL1}$ is first processed by a VSSM block and a $1\times1$ convolution, whereas $F_{GL2}$ is passed through a DConv module. 
To strengthen global-local coupling, we further process these intermediates, where $F_{GL1}$ is first processed by a VSSM block and a $1\times1$ convolution while $F_{GL2}$ is passed through a DW-Conv. 
%The resulting features are concatenated and further compressed via a $1\times1$ convolution to yield the 
The refined intermediate features are concatenated and passed through a $1\times1$ convolution to generate  final spatial representation $F_{Spa}$.
% This entire process facilitates progressive refinement of spatial features, where local details and global structures are jointly learned and fused.
% The entire computation process can be formulated as:
% \begin{equation}
% \begin{aligned}
% f_1, f_2, f_3, f_4 &= \mathcal{S}(F_{in}), \\
% F_{GL1} &= \mathcal{C}(\mathcal{P}(\mathcal{M}(f_1)), \mathcal{D}(f_2)),\\
% F_{GL2} &= \mathcal{C}(\mathcal{P}(\mathcal{M}(f_3)), \mathcal{D}(f_4)),          \\
% F_{Spa} &= \mathcal{P}(\mathcal{C}(\mathcal{P}(\mathcal{M}(F_{GL1})), \mathcal{D}(F_{GL2}))).          \\
% % F_{Spa} &= \mathcal{P}(\mathcal{M}(F_{GL3})).
% \end{aligned}
% \label{eq:formula1}
% \end{equation}
% Here, $\mathcal{S}(\cdot)$ denotes the split operation, $\mathcal{C}(\cdot)$ denotes concat, $\mathcal{P}(\cdot)$ denotes the pixel-wise convolution,  $\mathcal{M}(\cdot)$ denotes the VSSM block and $\mathcal{D}(\cdot)$ denotes the depth-wise convolution.
% This hierarchical design implements multi-stage global-local fusion, enabling Mamba to concurrently capture fine-grained textures and holistic spatial patterns. Crucially, iterative spatial interactions progressively enhance cross-context feature representations.
This hierarchical design enjoys these merits: \textit{i}) multi-stage fusion of global context and local details; \textit{ii}) concurrent capturing of fine-grained textures and holistic spatial patterns; \textit{iii}) progressively promoting the cross-context features through iterative refinement. 
% This structured design enforces a layered global-local fusion scheme, enabling Mamba to capture fine-grained texture while retaining holistic spatial understanding.
% Notably, this spatial-domain interaction procedure is applied twice sequentially, allowing the model to progressively enhance spatial semantics across both local and global contexts.

\subsubsection{Frequency-domain Branch.}
% Following the approach in~\cite{gao2024efficient}, the Frequency branch utilizes a 2D Fast Fourier Transform (2D-FFT) based architecture named FFCM to extract frequency-domain features $F_{fre}$ from the input scene.
Following~\cite{gao2024efficient}, our frequency branch employs a 2D Fast Fourier Transform (FFT)-based architecture called FFCM. This extracts frequency-domain features $F_{fre}$ from the input scene. 
% The frequency-domain branch is built upon the Fused Fourier Convolution Mixer (FFCM)~\cite{gao2024efficient}, which integrates Fast Fourier Transform (FFT) operations into a lightweight convolutional framework. 
By combining point-wise and multi-scale depth-wise convolutions, FFCM enhances its representational capacity while maintaining computational efficiency. 
% Projecting features into the frequency domain enables the module to capture global structural patterns and highlight the periodic characteristics of rain streaks, which are particularly salient in this domain.

\subsubsection{Dual-domain Fusion.}
To integrate the outputs of both the spatial and frequency branches, we concatenate them and apply a $1\times1$ convolution for channel compression. This fused representation is then added to the original input $F_{in}$ in a residual manner, producing the final output $F_{out}$. 
This design enables comprehensive scene feature extraction while maintaining computational efficiency.

%Finally, the spatial and frequency features are first concatenated and compressed via a pixel-wise convolutional layer to yield a fused representation, then this fused feature is added to the original input feature $F_{in}$ in a residual manner to produce the final output $F_{out} \in \mathbb{R}^{(b,c,h,w)}$. By integrating these two types of representations, the module enables comprehensive feature extraction across the scene. 

\subsection{Multi-modal Priors Guidance}
% The efficacy of multi-prior guidance in our framework critically depends on resolving inter-level feature conflicts.
% As illustrated in Fig~\ref{pipeline}, we address this through two synergistic components:
Effective multi-prior guidance requires resolving inter-level feature conflicts. 
As shown in Figure~\ref{pipeline}, we achieve this via two components:
\textit{i}) domain-specific feature adapters aligning prior-backbone representational spaces, and \textit{ii}) the Priors Fusion Injection (PFI) module alleviating conflicts during integration via structured attention.
% \textit{i}) domain-specific feature adapters that align the representational spaces between priors and backbone features, and \textit{ii}) the Prior Fusion Injection (PFI) module that actively resolves conflicts during multi-level prior integration through structured attention mechanisms.

\subsubsection{DINOv2 Adapter.}
The DINOv2 prior $P_{v}$ undergoes a three-stage adaptation for backbone integration: \textit{i}) initial channel reduction via pixel-wise convolution; \textit{ii}) spatial alignment via bilinear interpolation to match backbone resolutions; \textit{iii}) hierarchical representation generation using convolutional layers with PixelUnshuffle downsampling. 
This ensures dimensionally compatible prior injection across decoder stages while preserving structural semantics.

\begin{table*}[!htpb]
\caption{Quantitative comparison of PSNR/SSIM on five benchmark test datasets. Our MPHM achieves the best or comparable performance across all benchmarks. The best result for each dataset is highlighted in bold, and the second-best is underlined. }\vspace{-3mm}
%Metrics are reported in PSNR (dB) and SSIM.}
\centering
\small
\setlength{\tabcolsep}{2.5pt}  % 控制表格横向间距
\begin{adjustbox}{width=\linewidth}
\begin{tabular}{llcccccccccccccc}
\toprule
\multirow{2}{*}{\textbf{Type}} & \multirow{2}{*}{\textbf{Methods}} & 
\multicolumn{2}{c}{\textbf{Rain200H}} & \multicolumn{2}{c}{\textbf{Rain200L}} & 
\multicolumn{2}{c}{\textbf{DID-Data}} & \multicolumn{2}{c}{\textbf{DDN-Data}} & 
\multicolumn{2}{c}{\textbf{SPA-Data}} & \multicolumn{2}{c}{\textbf{Average}} \\

\cmidrule(lr){3-4} \cmidrule(lr){5-6} \cmidrule(lr){7-8} \cmidrule(lr){9-10} \cmidrule(lr){11-12} \cmidrule(lr){13-14}
& & PSNR & SSIM & PSNR & SSIM & PSNR & SSIM & PSNR & SSIM & PSNR & SSIM & PSNR & SSIM\\
\midrule
\multirow{2}{*}{Prior} 
& DSC~\cite{luo2015removing}   & 14.73 & 0.3815 & 27.16 & 0.8663 & 24.24 & 0.8279 & 27.31 & 0.8373 & 34.95 & 0.9416&25.67&0.7709 \\
& GMM~\cite{li2016rain}    & 14.50 & 0.4164 & 28.66 & 0.8652 & 25.81 & 0.8344 & 27.55 & 0.8479 & 34.30 & 0.9428&26.16& 0.7813\\
\midrule
\multirow{8}{*}{CNN} 
% & DDN~\cite{fu2017removing}        & 26.05 & 0.8056 & 34.68 & 0.9671 & 30.97 & 0.9116 & 30.00 & 0.9041 & 36.16 & 0.9457 &31.57&0.9068\\
& RESCAN~\cite{li2018recurrent}   & 26.75 & 0.8353 & 36.09 & 0.9697 & 33.38 & 0.9417 & 31.94 & 0.9345 & 38.11 & 0.9707 &33.25&0.9303\\
& PReNet~\cite{ren2019progressive}   & 29.04 & 0.8991 & 37.80 & 0.9814 & 33.17 & 0.9481 & 32.60 & 0.9459 & 40.16 & 0.9816&34.55&0.9512 \\
& MSPFN~\cite{jiang2020multi}     & 29.36 & 0.9034 & 38.58 & 0.9827 & 33.72 & 0.9550 & 32.99 & 0.9333 & 43.43 & 0.9843 &35.61&0.9517\\
& RCDNet~\cite{wang2020model}   & 30.24 & 0.9048 & 39.17 & 0.9885 & 34.08 & 0.9532 & 33.04 & 0.9472 & 43.36 & 0.9831 &35.97&0.9553\\
& MPRNet~\cite{zamir2021multi}  & 30.67 & 0.9110 & 39.47 & 0.9825 & 33.99 & 0.9590 & 33.10 & 0.9347 & 43.64 & 0.9844 &36.17&0.9543\\
& DualGCN~\cite{fu2021rain}  & 31.15 & 0.9125 & 40.73 & 0.9886 & 34.37 & 0.9620 & 33.01 & 0.9489 & 44.18 & 0.9902 &36.68&0.9604\\
& SPDNet~\cite{yi2021structure}    & 31.28 & 0.9207 & 40.50 & 0.9875 & 34.57 & 0.9560 & 33.15 & 0.9457 & 43.20 & 0.9871 &36.54&0.9594\\
& FADfomer~\cite{gao2024efficient}    & 32.48 & 0.9359 & 41.80 & 0.9906 & 35.48 & 0.9657 & 34.42 & 0.9602 & 49.21 & 0.9934 &38.67&0.9691\\
\midrule
\multirow{7}{*}{Transformer}
& Uformer~\cite{wang2022uformer}     & 30.80 & 0.9105 & 40.20 & 0.9860 & 35.02 & 0.9621 & 33.95 & 0.9545 & 46.13 & 0.9913 &37.22&0.9608\\
& Restormer~\cite{zamir2022restormer}  & 32.00 & 0.9329 & 40.99 & 0.9890 & 35.29 & 0.9641 & 34.20 & 0.9571 & 47.98 & 0.9921 &38.09&0.9670\\
& IDT~\cite{xiao2022image}              & 32.10 & 0.9344 & 40.74 & 0.9884 & 34.89 & 0.9623 & 33.84 & 0.9549 & 47.35 & 0.9930 &37.78&0.9666\\
& HCT-FFN~\cite{chen2023hybrid}  & 31.51 & 0.9100 & 39.70 & 0.9850 & 33.96 & 0.9592 & 33.00 & 0.9502 & 45.79 & 0.9898 &36.79&0.9588\\
& DRSformer~\cite{chen2023learning}  & 32.17 & 0.9326 & 41.23 & 0.9894 & 35.35 & 0.9646 & 34.35 & 0.9588 & 48.54 & 0.9924 &38.32&0.9675\\
& MSDT~\cite{chen2024rethinking}  & 32.45 & 0.9379 &  41.75 & 0.9904 & 35.37 & 0.9652 & 34.36 & 0.9593  & 49.07 & 0.9926 &38.60&0.9690\\
& NeRD-Rain~\cite{chen2024bidirectional}   & 32.40 & 0.9373 & 41.71 & 0.9903 & 35.53 & \underline{0.9659} & 34.45 & 0.9596 & 49.58 & 0.9940&38.73&0.9694\\
\midrule
\multirow{2}{*}{Mamba} 
&  TransMamba~\cite{sun2024hybrid}  & \underline{32.96} & \underline{0.9409} & 41.92 & \textbf{0.9938} & \underline{35.63} & 0.9657 & 34.72 & \underline{0.9603} & \underline{49.72} & \textbf{0.9968}&38.99&\underline{0.9715} \\
% & FreqMamba~\cite{zou2024freqmamba}     & -- & -- & -- & -- & -- & -- & -- & -- & -- & --&& \\
& FourierMamba~\cite{li2024fouriermamba}  & 32.71 & 0.9395 & \underline{42.27} & 0.9908 & 35.49 & \underline{0.9659} & \textbf{35.58} & 0.9599 & 49.18 & 0.9931&\underline{39.04}& 0.9698\\
\midrule
&\textbf{MPHM(Ours)} & \textbf{33.53} & \textbf{0.9475} & \textbf{42.30} & \underline{0.9913} & \textbf{35.65} & \textbf{0.9662} & \underline{34.80} & \textbf{0.9607} & \textbf{49.76} & \underline{0.9951}&\textbf{39.21}& \textbf{0.9722}\\
\bottomrule
\end{tabular}%\vspace{-4mm}

\label{tab:sota_comparison}\vspace{-6mm}

\end{adjustbox}
\end{table*}

\subsubsection{CLIP Adapter.}
% A lightweight bottleneck adapter adapts the CLIP text prior $P_{t}$ for task-specific integration. 
A lightweight bottleneck adapter processes the CLIP text prior: \textit{i}) using linear projection to reduce dimensionality to a latent space; \textit{ii}) employing non-linear layers to extract task-specific semantics; \textit{iii}) aligning features to visual representations.
%non-linear processing, and dimensional restoration, creating an information filter that retains task-critical semantics while optimizing alignment with visual representations.

\subsubsection{Priors Fusion Injection.}
The adapted priors are integrated through a two-stage attention mechanism.
Using DINOv2's scene-aware information $P_{v}$ as keys/values provides cues for structural semantic retrieval. Employing CLIP-derived features $P_{t}$
as queries focuses on task-relevant regions. 
%for task-relevant spatial attention. 
This sequential strategy prevents fusion conflicts. A subsequent multi-head self-attention module captures long-range dependencies. Finally, a Gated Depth-wise Feedforward Network (GDFN)~\cite{zamir2022restormer} refines the features via its dual-gating mechanism, balancing transformation and spatial preservation.
% This sequential approach delivers complementary guidance while preventing fusion conflicts. 
% Building on this foundation, a multi-head self-attention module then captures long-range dependencies across the entire feature map. Finally, the representation undergoes refinement through a Gated Depth-wise Feedforward Network (GDFN)~\cite{zamir2022restormer}, which adaptively balances feature transformation with spatial preservation using its dual-gating mechanism.

\section{Experiments}
% This section details the implementation and training protocols of our MPHM framework, followed by comprehensive experimental validation. We rigorously evaluate the method's effectiveness through extensive benchmarking on both synthetic and real-world datasets, employing multiple metrics to compare against state-of-the-art deraining approaches.
% We present MPHM’s implementation and training protocols, accompanied by a comprehensive evaluation. Rigorous benchmarking is conducted against 19 state-of-the-art deraining methods from the last decade, leveraging multiple metrics on both synthetic and real-world datasets.
% We detail MPHM’s implementation and training protocols and comprehensively evaluate its effectiveness. To ensure rigorous benchmarking, we compare MPHM against 19 state-of-the-art deraining methods proposed over the past decade, using multiple metrics on both synthetic and real-world datasets.
We detail MPHM's implementation and training protocols and then comprehensively evaluate its effectiveness. We compare against 19 state-of-the-art deraining methods over the last decade on synthetic and real-world datasets with multiple metrics. 

\subsection{Implementation Details}
\subsubsection{Datasets.}
The proposed method is evaluated on five benchmark datasets: Rain200L/H ~\cite{yang2017deep}, DID-Data~\cite{zhang2018density}, DDN-Data~\cite{fu2017removing}, and SPA-Data~\cite{wang2019spatial}, involving various rain patterns and scene complexities. 
The Rain200L dataset comprises 1,800 synthetically generated rainy/clean image pairs for training, along with a separate test set of 200 image pairs. 
As its counterpart, Rain200H maintains identical data partitioning while exclusively containing heavy rain conditions.
The DID-Data provides 12,000 synthetic training pairs accompanied by 1,200 dedicated test pairs, whereas DDN-Data contains 12,600 training pairs with 1,400 test samples.
SPA-Data is a large-scale real-world benchmark that contains 638,492 training pairs and a standardized test set of 1,000 image pairs with ground truth.
Furthermore, we evaluate the model's generalization capability on two real-world datasets: Internet-Data~\cite{wang2019spatial} and RE-RAIN~\cite{chen2023towards}, containing 147 and 300 rainy images without ground truth, respectively.

\subsubsection{Metrics.}
For benchmark datasets with ground truth, we report PSNR~\cite{huynh2008scope} and SSIM~\cite{wang2004image} for evaluation. For unpaired real-world data, we use no-reference metrics: BRISQUE~\cite{mittal2012making} and NIQE~\cite{mittal2012no} to evaluate perceptual quality. 
% Following established practices, we utilize PSNR~\cite{huynh2008scope} and SSIM~\cite{wang2004image} quantitative metrics for benchmark datasets with available ground truth. 
% For unpaired real-world data, no-reference quality assessment metrics, including BRISQUE~\cite{mittal2012making}, NIQE~\cite{mittal2012no}, and SSEQ~\cite{liu2014no}, are employed to evaluate perceptual quality under uncontrolled conditions.

\subsubsection{Experimental Settings.} %Implementation Details.}
% The depth of the HMM module in each stage is set to $\{4, 6, 8, 6, 4\}$, and the initial channel dimension $C$ is set to $32$. 
% The network is trained with Adam optimizer~\cite{loshchilov2017fixing}, where the learning rate is initialized at $1 \times 10^{-3}$ and gradually decayed to $1 \times 10^{-5}$ following a cosine annealing strategy~\cite{loshchilov2016sgdr}.
% The training samples are %center-cropped to 
% centered on a resolution of $256 \times 256$ and the batch size is set to $4$. All experiments are conducted using the PyTorch~\cite{paszke2019pytorch} framework on NVIDIA 3090 GPUs.
The HMM depth per stage is $\{4, 6, 8, 6, 4\}$ with an initial channel dimension $C = 32$. 
Model training employs the Adam optimizer~\cite{loshchilov2017fixing} with a $1 \times 10^{-3}$ initial learning rate decayed to $1 \times 10^{-5}$ via cosine annealing~\cite{loshchilov2016sgdr}. The inputs are center-cropped to $256 \times 256$ resolution with a batch size of $4$. All experiments run on PyTorch~\cite{paszke2019pytorch} using NVIDIA 3090 GPUs.

\begin{figure*}[!htpb]
  \centering
  \includegraphics[width=\textwidth]{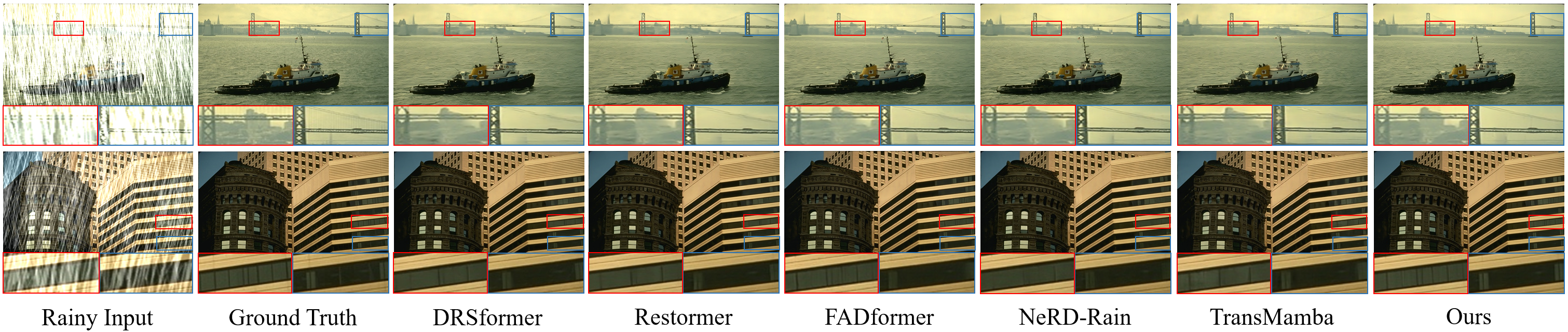}\vspace{-2mm}
  \caption{Visual results on Rain200H. Our method recovers clearer details and textures.
  %Derained results on the Rain200H dataset. Compared with the derained results, our method recovers a high-quality image with clearer details. 
 Please zoom in for a better view. 
  %the figures offers a better view at the deraining capability. 
  }\vspace{-2mm}
  \label{Compare_200H}%\vspace{-4mm}
\end{figure*}

\begin{figure*}[htbp]
  \centering
  \includegraphics[width=\linewidth]{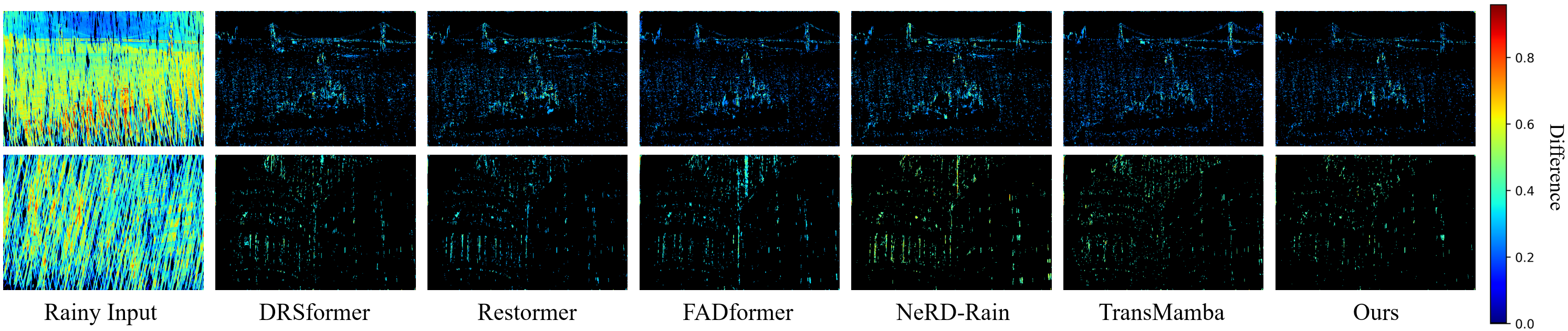}\vspace{-2mm}
  \caption{Residual heatmaps between derained results and ground truth. Brighter colors denote larger pixel-wise deviations. Our method shows minimal differences.
  %Residual maps between derained results and ground-truth. Brighter colors indicate larger pixel-wise differences, where our method achieves the most consistent and minimal differences.
  }\vspace{-2mm}
  \label{res_hot_map}
\end{figure*}

\begin{table*}[!htb]
\caption{Quantitative evaluations on the unpaired RE-RAIN dataset. We report BRISQUE ($\downarrow$) and NIQE ($\downarrow$), no-reference image quality metrics where lower values indicate better perceptual quality. Our method achieves the best scores.}\vspace{-2mm}
\centering
\small
\setlength{\tabcolsep}{2.3pt}
\begin{tabular}{lccccccccccc}
\toprule
\textbf{Methods} & \textbf{Rainy Input}  &  \textbf{HCT-FFN}& \textbf{Restormer}  & \textbf{IDT}  & \textbf{DRSformer} & \textbf{NeRD-Rain} & \textbf{FADformer}& \textbf{TransMamba}& \textbf{Ours}\\
\midrule
BRISQUE \ $\downarrow$& 25.279 &35.425&28.410   & 23.704  & 22.857& 21.693& \underline{21.670}&23.420&\textbf{21.221}\\
% Model-2 & \checkmark  &\checkmark& $\times$&  &  &  &  \\
NIQE \ $\downarrow$ & 4.169  &4.938 & 4.106& 3.918 &3.875 & 3.862 &3.980&\underline{3.844}&\textbf{3.787}\\

\bottomrule
\end{tabular}

\label{tab:unpaired}\vspace{-4mm}
\end{table*}

\subsection{Comparison with State-of-the-arts}

\subsubsection{Synthetic Datasets Results.}
% Quantitative results on four synthetic datasets are presented in Table~\ref{tab:sota_comparison}. Compared with methods proposed in the past decade within this field, our proposed MPHM consistently achieves the best performance with an average improvement of 0.15 dB.
Quantitative results in four synthetic datasets are shown in Table~\ref{tab:sota_comparison}, demonstrating that our MPHM outperforms existing approaches with an obvious performance gain. %by achieving an average gain of 0.17 dB PSNR. 
In the Rain200H and Rain200L datasets, MPHM attains a cumulative improvement of 0.85 dB, showing superior generalization in degradation complexities. 
Figure~\ref{Compare_200H} highlights the ability of MPHM to enhance texture preservation, including scenarios when background patterns resemble rain streaks (\emph{e.g.}, oblique or vertical lines). 
Figure~\ref{res_hot_map} visualizes the pixel-wise residuals. It is obvious that our method yields sparser, lower-intensity residuals (darker heatmaps), confirming effective rain removal and detail preservation.
This considerable superiority comes from semantic information supplementation and confusion elimination, where conventional methods mistakenly remove object structures alongside rain artifacts. 
Our semantic prior injection enables precise residual-background separation, maximizing detail retention in heavy rain.

\begin{figure*}[!htpb]
  \centering
  \includegraphics[width=\textwidth]{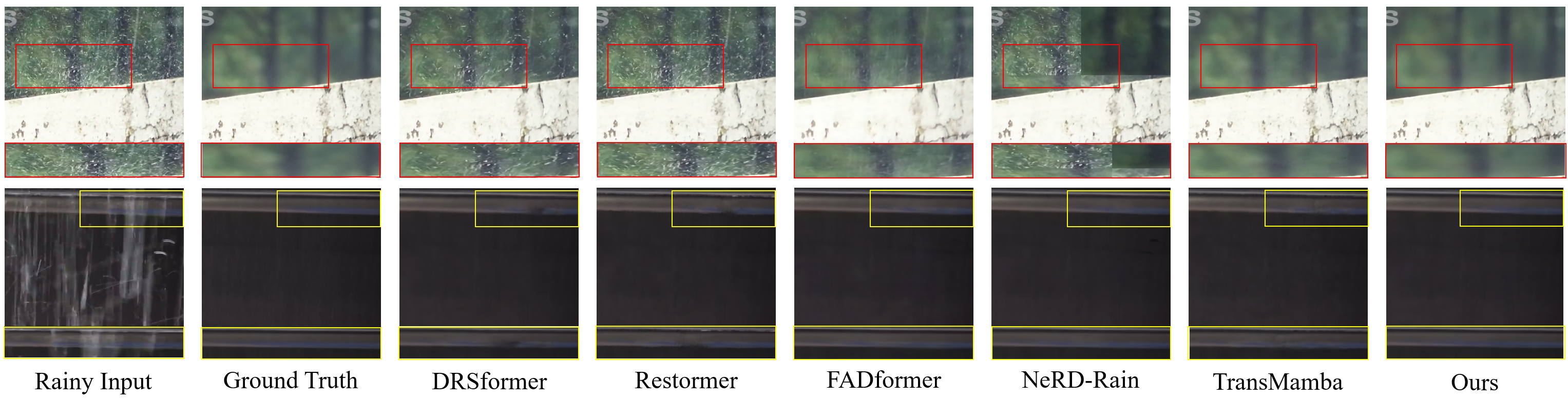}\vspace{-2mm}
  \caption{Derained results on the  real-world paired SPA-Data. Compared with the derained results, our approach achieves superior rain removal performance and simultaneously preserves structural details. Please zoom in for a better view.}\vspace{-2mm}
  \label{Spa_vis}%\vspace{-4mm}
\end{figure*}

\subsubsection{Real-world Datasets Results.}
\begin{figure*}[htbp]
  \centering
  \includegraphics[width=\linewidth]{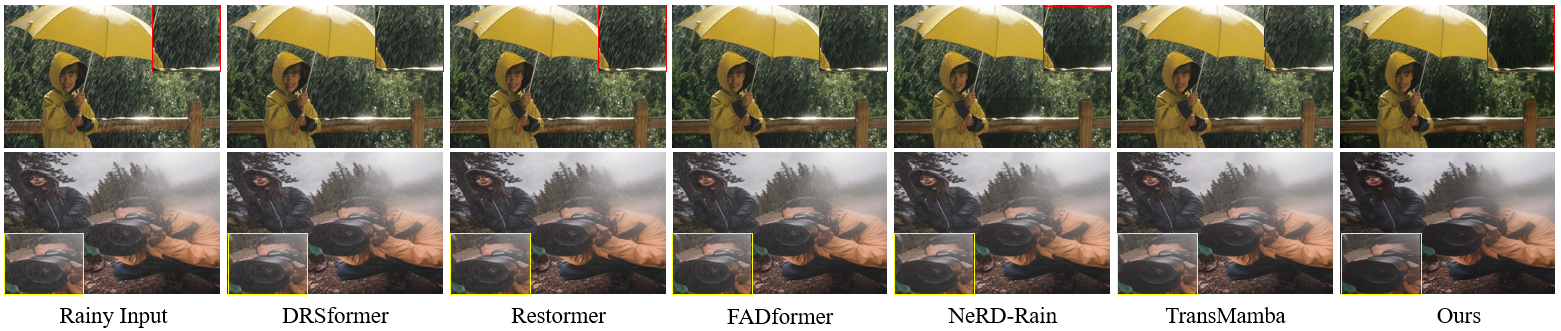}\vspace{-2mm}
  \caption{Visualizing on real-world unpaired RE-RAIN (top) and Internet-Data (bottom).}
  \label{Rerain_internet}\vspace{-2mm}
\end{figure*}

We further validate our approach on the real-world SPA-Data benchmark. 
As shown in Table~\ref{tab:sota_comparison}, our MPHM achieves the best performance in all indicators. Figure~\ref{Spa_vis} shows that the derained images produced by our MPHM %confirms improved
enjoy better texture preservation and less rain residue. Table~\ref{tab:unpaired} tabulates the comparison of cross-domain generalization, where models trained on the Rain200H dataset are tested on unpaired RE-RAIN data. %As shown in Table~\ref{tab:unpaired}, 
As expected, our method achieves the lowest BRISQUE and NIQE scores, indicating superior perceptual quality. Figure~\ref{Rerain_internet} further validates the robust capability of rain removal and detail reconstruction in real-world scenarios, bridging the synthetic-to-real domain gap.
% As shown in the rightmost column of Table~\ref{tab:sota_comparison}, MPHM achieves the best-performing baseline.
% Qualitative results in Figure~\ref{Spa_vis} demonstrate superior visual quality, with enhanced texture preservation and rain removal attributable to our method’s design. 
% To evaluate generalization capability, we directly test models trained on Rain200H using the unpaired RE-RAIN dataset. 
% As shown in Table~\ref{tab:unpaired}, our method outperforms all competitors, achieving the lowest BRISQUE and NIQE scores, which indicates superior perceptual quality and naturalness.
% Figure~\ref{Rerain_internet} further confirms that our model removes rain more thoroughly while reconstructing fine details with high realism under diverse real-world conditions. 
% These results collectively demonstrate our design’s effectiveness in bridging the synthetic-to-real domain gap.

\begin{table}[t]
\centering
\small

\setlength{\tabcolsep}{6pt}
\caption{Comparison of model complexity and performance against state-of-the-art methods. The size of the test image is $256\times256$ pixels. ``FLOPs" (in G) and ``Params" (in M) denote the floating-point operations and the number of trainable parameters, respectively.}\vspace{-2mm}
\label{tab:complex}
\begin{adjustbox}{width=\linewidth}
\begin{tabular}{lcccc}
\toprule
\textbf{Methods}   & Restormer& IDT& DRSformer&FADformer  \\
\midrule
\textbf{FLOPs (G)}      & 174.7 & 61.90& 220.33&48.51\\
\textbf{Par. (M)}    & 26.12& 16.41& 33.65 &\textbf{6.96}\\
\textbf{PSNR. (db)}    & 38.09  & 37.78 &38.32 &38.67  \\
\midrule
\textbf{Methods}   & NeRD-Rain & TransMamba& FourierMamba & Ours \\
\midrule
\textbf{FLOPs (G)}     & 148.05  & \underline{45.67} & \textbf{22.56}& 61.89 \\
\textbf{Par. (M)}    & 22.89  & 16.74 &17.62 &\underline{10.28}  \\
\textbf{PSNR. (db)}    & 38.73  & 38.99 &\underline{39.04} &\textbf{39.21}  \\
\bottomrule
\end{tabular}
\end{adjustbox}
\end{table}

\subsubsection{Model Complexity.}
To evaluate the model efficiency, we measure parameters and FLOPs using $256\times256$ input. 
As summarized in Table~\ref{tab:complex}, MPHM maintains a favorable computational profile while balancing deraining performance and inference efficiency. %achieving strong restoration quality while remaining efficient across both metrics. 

\subsection{Ablation Studies}
Ablation studies are conducted on the Rain200H with consistent training configurations to allow fair performance comparison between variants.

\begin{table}[t]
\caption{Component-level ablations of the proposed HMM. The results validate the effectiveness of frequency and spatial encoding.}\vspace{-2mm}
\centering
\small
\setlength{\tabcolsep}{2.1pt}
\begin{tabular}{lccccccc}
\toprule
\textbf{Variant} & \textbf{FFCM}  & \textbf{DW}& \textbf{VSSM}& \textbf{PSNR}  & \textbf{SSIM}  & \textbf{Params} & \textbf{FLOPs} \\
\midrule
Model-1& $\times$  & \checkmark&\checkmark&  30.56 &0.9158  &2.307  & 23.54 \\
% Model-2 & \checkmark  &\checkmark& $\times$&  &  &  &  \\
Model-2 & \checkmark & $\times$ &\checkmark & 32.93& 0.9407& 8.470 & 69.58 \\
Ours & \checkmark & \checkmark & \checkmark&33.06 & 0.9421 & 7.247 & 57.76 \\
\bottomrule
\end{tabular}\vspace{-3mm}
\label{tab:HMM}
\end{table}
\subsubsection{Effectiveness of HMM.}
\begin{figure}[htbp]
  \centering
  \includegraphics[width=\linewidth]{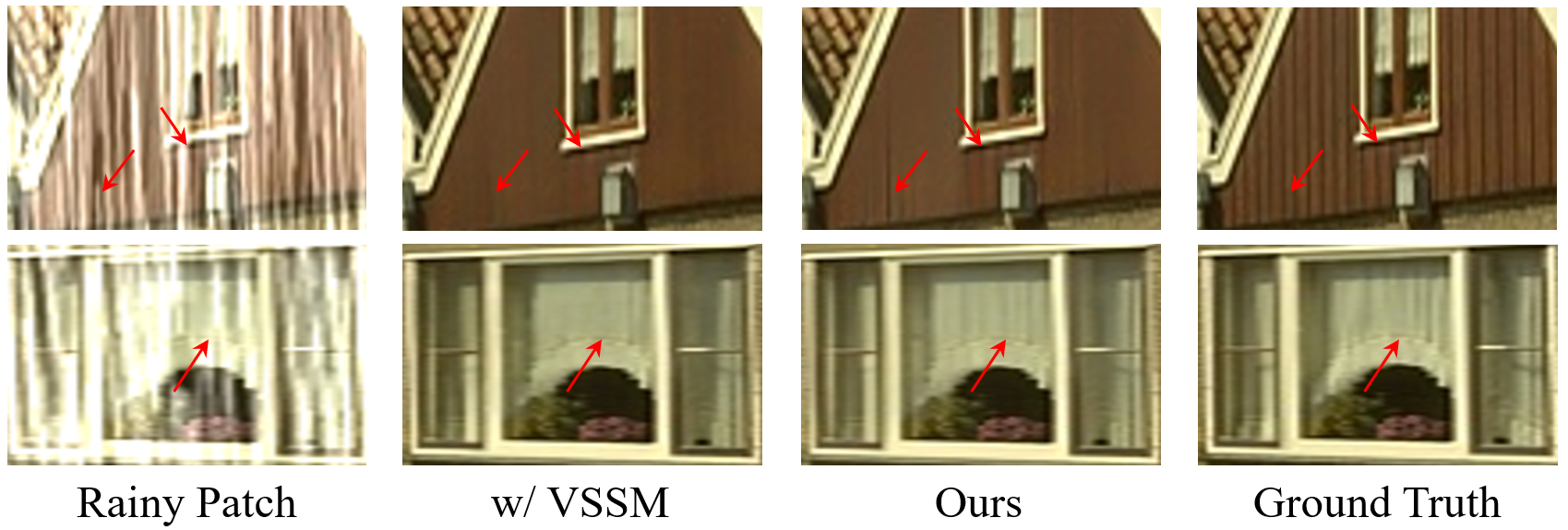}
  \caption{Qualitative comparison of local detail recovery. Our strategy enhances the VSSM’s capacity to reconstruct spatial structures.
  %Qualitative comparison of local spatial detail recovery. Our grouped local-global strategy  effectively enhances the original VSSM’s capability in modeling local spatial structures.
  }
  \label{fig:aba_hmm}
\end{figure}

As shown in Table~\ref{tab:HMM}, the results demonstrate the critical contributions of the HMM components. 
Removing the frequency-domain branch (FFCM) in Model-1 causes a 2.50 dB drop in PSNR, confirming that frequency information is essential for preserving structural patterns and suppressing residual rain streaks. 
Model-2 shows a 0.13 dB PSNR reduction compared to the full HMM while requiring 16.9\% more parameters and 20.5\% higher FLOPs, indicating suboptimal efficiency. Figure~\ref{fig:aba_hmm} further reveals Model-2’s diminished local spatial awareness (manifested as blurred high-frequency textures) due to the omitted feature grouping and DW-Conv.
Conversely, our grouped local-global strategy combines parallel DW-Conv (local refinement) with VSSM (global context), reducing complexity while enhancing local modeling for superior efficiency and fidelity.
% In contrast, our grouped local-global strategy harmonizes merits of parallel DW-Conv (local refinement) and Vision State Space Model (VSSM) (global context), simultaneously reducing complexity and enhancing local feature modeling for superior efficiency and fidelity.
% To assess component effectiveness in the proposed Hybrid Mamba Module (HMM), ablation studies (Table~\ref{tab:HMM}) reveal critical insights. Removing the frequency-domain branch (FFCM) in Model-1 causes a sharp PSNR drop of 2.39 dB, confirming frequency information's indispensable role in preserving structural patterns and mitigating residual rain streaks.
% Model-2 exhibits a 0.13dB PSNR drop relative to the full HMM while incurring 16.9\% more parameters and 20.5\% higher FLOPs, underscoring its inferior performance and computational inefficiency.
% Crucially, Figure~\ref{fig:aba_hmm} reveals Model-2's compromised local spatial awareness—manifested as blurred textures in high-frequency regions—due to omitted feature grouping and DW-Conv. 
% In contrast, our grouped local-global strategy leverages parallel DW-Conv (local refinement) and VSSM (global context) to simultaneously reduce complexity and enhance local modeling capacity, achieving both efficiency and fidelity gains.

\subsubsection{Ablation on Branch Fusion Scheme.}
We evaluate three fusion strategies for integrating frequency and spatial branches in the HMM module: direct addition, cross-attention, and our concatenation-based approach with a $1\times1$ convolution.
As shown in Table~\ref{tab:branch_fsuion}, our method achieves the highest PSNR/SSIM with moderate complexity. 
Direct addition yields slightly inferior performance due to %inadequate cross-channel feature interaction
feature redundancy and interference. Cross-attention incurs the highest computational cost and the worst restoration quality, suggesting over-parameterization. Our lightweight concatenation method thus optimally balances restoration efficacy and efficiency.

\begin{table}[t]
\centering
\small
\setlength{\tabcolsep}{2.5pt}
\caption{Ablation studies for branch fusion scheme.}\vspace{-2mm}
\label{tab:branch_fsuion}
\begin{tabular}{lccccc}
\toprule
\textbf{Strategy} &  Addition & Cross-attn & Concat~(Ours) \\
\midrule
\textbf{PSNR/SSIM} &32.82/0.9397 &32.22/0.9339& 33.06/0.9421  \\
\textbf{Par./FLOPs} &6.864/53.74 &7.352/ 59.31& 7.247/57.76\\
\bottomrule
\end{tabular}\vspace{-3mm}
% \end{adjustbox}
\end{table}

% Model-3 & \checkmark & $\times$& $\times$ &\checkmark &32.44 & 0.9347 & 7.898 & 63.94 \\

% \subsubsection{Effectiveness of priors adapters.}
% Quantitative ablation (Table~\ref{tab:adapter}) rigorously validates adapter necessity. 
% Removing DINOv2's visual adapter (Config-A) causes severe performance degradation, confirming structural semantics loss. 
% Without CLIP's textual adapter (Config-B) retains higher PSNR but exhibits significant SSIM drop, indicating residual artifact persistence.
% Our full adapter implementation achieves optimal performance by establishing precise feature-space alignment between foundation model priors and deraining backbone. 
% The adapters enable seamless injection of complementary guidance—DINOv2’s structural semantics and CLIP’s task-aware signals—through domain-adapted representation transformation with minimal overhead.
% \begin{table}[t]
% \caption{Ablation studies for priors adapters.}
% \footnotesize
% \centering
% % \small
% \setlength{\tabcolsep}{2.3pt}
% \begin{tabular}{lcccccc}
% \toprule
% \textbf{Variant} & \textbf{Ada-D}  & \textbf{Ada-C}& \textbf{PSNR}  & \textbf{SSIM}  & \textbf{Params} & \textbf{FLOPs} \\
% \midrule
% Config-A& $\times$  & \checkmark&&  &   &  \\
% Config-B & \checkmark  &$\times$&    && 10.024 &  62.16\\
% Ours & \checkmark & \checkmark &33.22 & 0.9464 & 10.041 & 62.49 \\
% \bottomrule
% \end{tabular}

% \label{tab:adapter}
% \end{table}
\begin{figure*}[htbp]
  \centering
  \includegraphics[width=\linewidth]{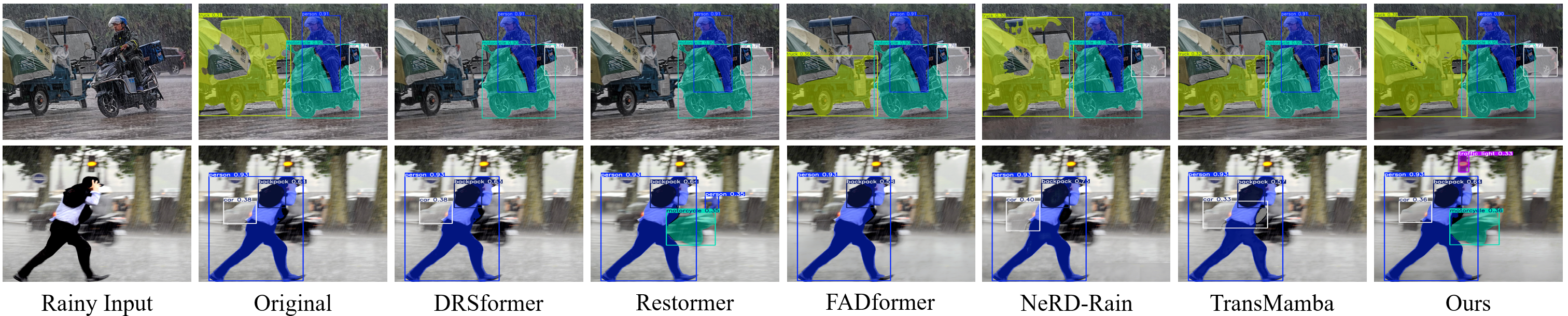}\vspace{-2mm}
  \caption{Visual comparison of object detection and instance segmentation enhanced by different deraining methods. Our method preserves clearer structure and yields more accurate visual perception. Please zoom in for a better view.}
  \label{down_stream_task}\vspace{-2mm}
\end{figure*}

\subsubsection{Ablation on Priors Injection.}
\begin{figure}[htbp]
  \centering
  \includegraphics[width=\linewidth]{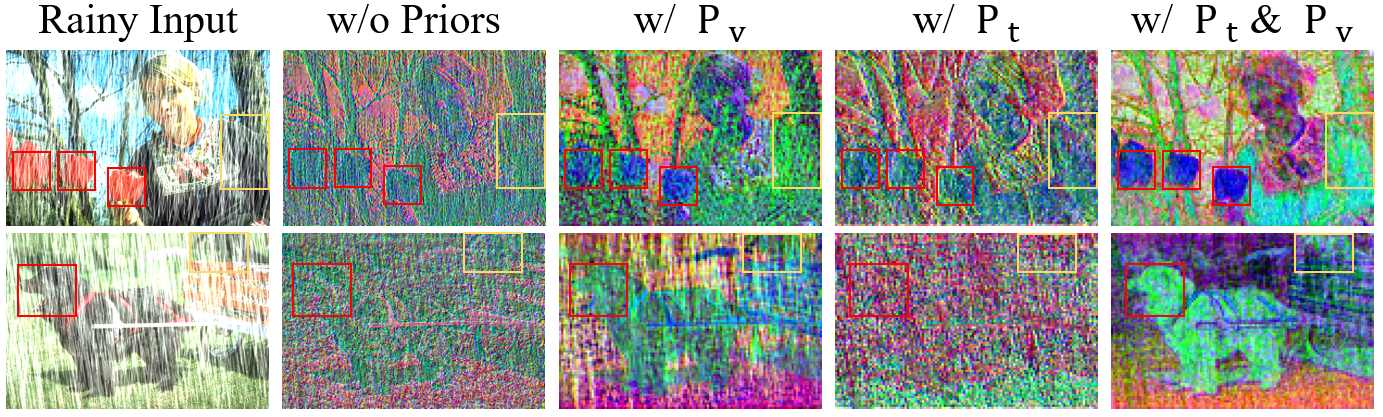}\vspace{-2mm}
  \caption{PCA visualization of different prior features from the bottleneck layer. %PCA visualization of bottleneck representations under different prior injection settings.
  }\vspace{-2mm}
  \label{pfi_aba}
\end{figure}

We ablate four prior-guided representation learning configurations, involving \textit{no priors}, \textit{visual prior} ($P_v$) only, \textit{textual prior} ($P_t$) only, and joint injection (Table~\ref{tab:PFI}). The baseline without priors achieves only 33.06 dB PSNR and 0.9421 SSIM, exhibiting noisy, entangled features that confuse rain streaks from scene contents (Figure~\ref{pfi_aba}). 
Introducing $P_v$ (from DINOv2) moderately improves structural recovery but overemphasizes rain streaks that resemble scene patterns. Using $P_t$ alone enhances rain suppression, but produces over-smooth textures. 
Joint injection of both priors maximizes performance by combining $P_v$’s structural awareness with $P_t$’s task-specific guidance, enabling robust semantic disentanglement and visually consistent deraining (Figure~\ref{pfi_aba}).

\begin{table}[t]
\caption{Ablation studies on priors injection in PFI. ``$P_v$'' and ``$P_t$'' indicate the visual and textual priors, respectively.}\vspace{-2mm}
\centering
\small
\setlength{\tabcolsep}{2.5pt}
\begin{tabular}{lcccc}
\toprule
\textbf{Variant} & \textbf{P\textsubscript{v}} & \textbf{P\textsubscript{t}} & \textbf{PSNR}  & \textbf{SSIM}   \\
\midrule
$w/o \ Priors$ & $\times$ & $\times$  & 33.06 & 0.9421  \\
$w \ P_{v}$ & \checkmark & $\times$  & 33.22 & 0.9446  \\
$w \ P_{t}$ & $\times$ & \checkmark  & 33.20 & 0.9434  \\
\textbf{$w \ P_{v}\&\ P_{t}(Ours)$} & \checkmark & \checkmark & 33.53 & 0.9475  \\
\bottomrule
\end{tabular}

\label{tab:PFI}\vspace{-2mm}
\end{table}

\subsubsection{Ablation on Priors Fusion Scheme.}
We ablate four fusion strategies to integrate visual ($P_v$) and textual ($P_t$) priors, involving addition, concatenation, cross-attention, and our hierarchical fusion. 
As shown in Table~\ref{tab:priors_fsuion}, both the addition and concatenation schemes degrade restoration performance due to non-adaptive merging that equally weights features irrespective of semantic relevance.
This propagates conflicting errors, making it difficult to learn semantic relation representations across modalities.
Although the cross-attention scheme brings marginal gains, single-stage fusion forces incompatible merging of $P_v$’s micro scene abstractions and $P_t$’s macro task guidance. 
Our hierarchical fusion overcomes these limitations by decoupling injection: $P_v$ adapts scene representations in early stages, while $P_t$ refines task-specific features later. This progressive reconciliation eliminates semantic conflicts, enabling substantial semantic fusion and robust feature representation.

\begin{table}[t]
\centering
\small
\setlength{\tabcolsep}{1.8pt}
\caption{Ablation studies on priors fusion scheme.}\vspace{-2mm}
\label{tab:priors_fsuion}
\begin{adjustbox}{width=\linewidth}
\begin{tabular}{lcccccc}
\toprule
\textbf{Strategy} & Addition & Concat & Cross-attn& Ours \\
\midrule
\textbf{PSNR/SSIM} &31.95/0.9287 &32.05/0.9324& 33.15/0.
9448&33.53/0.9475  \\
% \textbf{Par./FLOPs} &-/- &-/-&-/- &10.28/61.89\\
\bottomrule
\end{tabular}\vspace{-2mm}
\end{adjustbox}
\end{table}

\subsection{Impact on Downstream Task}
% To validate real-world applicability, we deploy derained images in YOLOv8 for object detection and instance segmentation on the RE-RAIN dataset. Figure~\ref{down_stream_task} confirms significant performance gains over state-of-the-art methods. Our multi-prior strategy enhances semantic understanding, preserving object structures under heavy rain through adaptive scene reasoning. Simultaneously, the HMM module excels in detail reconstruction, recovering critical high-frequency textures essential for contour recognition. This synergy of semantic comprehension and detail fidelity demonstrates robust utility for adverse-weather vision systems, with consistent improvements across both detection and segmentation metrics.
To rigorously validate real-world utility, we evaluate derained images using YOLOv8~\cite{UltralyticsYOLOv8} for object detection and instance segmentation on the RE-RAIN dataset. As shown in Figure~\ref{down_stream_task}, our approach achieves substantial performance gains over state-of-the-art methods (better detection segmentation accuracy). These improvements directly stem from our dual technical innovations. i) The multi-prior fusion strategy (integrating DINOv2's pixel-level visual cues and CLIP's contextual language priors) enables adaptive scene representation, preserving structural coherence of objects under heavy occlusion. This resolves semantic ambiguity in degraded regions (\emph{e.g.}, distinguishing rain-obscured cars from background clutter). ii) The Hierarchical Mamba Module (HMM) facilitates multi-scale feature refinement through global-local interactions in both spatial and frequency domains. By recovering high-frequency textures (critical edges and contours), HMM overcomes Mamba’s inherent local detail limitations, enabling precise object boundary delineation. The synergy of semantic preservation and textural fidelity underpins MPHM’s robustness for adverse-weather vision systems. Quantitative gains in detection/segmentation metrics confirm our method’s superiority in maintaining task-critical information. 

\section{Conclusion}
In this work, we proposed MPHM, a multi-prior hierarchical Mamba framework for single image deraining. 
By strategically combining macro-level textual priors from CLIP with micro-level visual priors from DINOv2, MPHM delivers comprehensive semantic guidance that improves discrimination between rain streaks and background structures. 
Furthermore, our Hierarchical Mamba Module (HMM) introduces a global-local interaction mechanism and Fourier-enhanced representations, effectively addressing the spatial locality limitations of conventional Mamba architectures. 
Extensive experiments on synthetic and real-world datasets demonstrate that MPHM achieves state-of-the-art performance while maintaining competitive model complexity. 
Future work will explore extending our multi-prior fusion paradigm to handle co-occurring degradations such as haze and illumination imbalance, further improving robustness under diverse weather conditions.

\section{Acknowledgments}
This research was financially supported by the National Natural Science Foundation of China (62501189, U23B2009), the Natural Science Foundation of Heilongjiang Province of China for Excellent Youth Project (YQ2024F006).
\bibliography{aaai2026}

% \bibliography{yourbib}
% \clearpage
% \input{ReproducibilityChecklist}

\end{document}